# Automated Agriculture Commodity Price Prediction System with Machine Learning Techniques

Zhiyuan Chen[*,1], Howe Seng Goh[1], Kai Ling Sin[1], Kelly Lim[1], Nicole Ka Hei Chung[1], Xin Yu Liew[1]

[1]School of Computer Science, University of Nottingham Malaysia, Semenyih, 43500, Malaysia

A B S T R A C T

*The intention of this research is to study and design an automated agriculture commodity price prediction system with novel machine learning techniques. Due to the increasing large amounts historical data of agricultural commodity prices and the need of performing accurate prediction of price fluctuations, the solution has largely shifted from statistical methods to machine learning area. However, the selection of proper set from historical data for forecasting still has limited consideration. On the other hand, when implementing machine learning techniques, finding a suitable model with optimal parameters for global solution, nonlinearity and avoiding curse of dimensionality are still biggest challenges, therefore machine learning strategies study are needed. In this research, we propose a web-based automated system to predict agriculture commodity price. In the two series experiments, five popular machine learning algorithms, ARIMA, SVR, Prophet, XGBoost and LSTM have been compared with large historical datasets in Malaysia and the most optimal algorithm, LSTM model with an average of 0.304 mean-square error has been selected as the prediction engine of the proposed system.*

## 1. Introduction

The increasing availability of large amounts of agricultural commodity prices historical data and the need of performing accurate predicting of price fluctuations in agricultural economy demands the definition of robust and efficient techniques able to infer from current observations. The traditional way to solve the prediction problem lies in linear statistical methods (such as ARIMA models), and more recently with the emergence of machine learning techniques, the solution has largely shifted from statistical methods to machine learning area. However, the selection of proper set from historical data for forecasting still has limited consideration. On the other hand, when implementing machine learning techniques, finding optimal parameters of learning algorithm for global solution, nonlinearity and avoiding curse of dimensionality are still biggest challenges, therefore machine learning strategies study are needed.

In practice, volatility in price of agricultural commodities is often unpredictable as they are affected by eventualities for example fluctuations of oil price, greenhouse effects and natural disasters such as flood or attacks by disease. Uncertainties of agricultural commodities price endanger the accessibility of food by consumers which leads to food insecurity and causes starvation and malnutrition. Instability of agricultural commodities price due to oversupply or lack of demand causes unnecessary food wastage. In recent years, the price of agricultural commodities such as for palm oil and rubber have been steadily decreased. This downward trend is making an impact to Malaysia's economy and can contribute to a slower economic growth in various investments in Malaysia. There are many exogenous factors that could cause such a trend and time-series data analysis is required to forecast such trends to improve Malaysia's agricultural plantation plan for better country development. Being able to anticipate the fluctuations and patterns in agricultural commodities price will enable the government to propose new policies that can help prevent the country into worse economy state. Further, the agricultural commodities providers are able to control their supply based on the time series analysis in order to prevent a bad plantation plan.

Most of the time series analysis on agricultural commodities is based on the US and China commodities market and there is no concrete research done on the Malaysia commodities market. Thorough agricultural commodities prices analysis should be performed based on the Malaysia prices.

---

*H. David et al. / Advances in Science, Technology and Engineering Systems Journal Vol. 6, No. 2, XX-YY (2021)*

This paper is an extension of work originally presented in ITCC 2020 [1].

## 2. Background

Considering the earliest studies, some analysis on commodity futures markets have been done. William G et al. discovered that futures prices are good price predictors in the corn and soybean market established from empirical forecast assessment [2]. There are many other subsequent studies on different agricultural futures markets in 1970's and it is the result of high dependency used by the earliest researchers which is focusing on specific market conditions and traditional econometric models [3]. Looking in the traditional econometric model used during the earliest days, there are two existing problems in literature regarding the agricultural future prediction. First is considering the limited analysis due to the capacity of the model and high computational cost have led to high dimensional multivariate time series prediction [4] [5]. Second is based on those models that are estimated through assumption that variables are independent, normal distribution, which is unrealistic in the real world[6].

Later on several research studies have proposed to implement agriculture price prediction scheme using different machine learning algorithms [7] [8] [9] [10]. However, the performance of machine learning in agricultural commodity futures prediction is rarely explored. Future forecasting is usually done by analysing past price data of each commodity, climate, location, planting area, and several other conditions. Therefore, it is challenging because of its inherent complexity and dynamism. According to the lowest error percentage, many have selected ANN and PLS as prediction algorithms [8]. In a research paper by Arun Kumar proposed a system to apply prediction by analyzing past soil and rainfall datasets [9]. Apart from that, another research paper from Askunuri Manjula has done crop prediction using multiple features, such as weather forecasting, pesticides and fertilizers and past revenue. Some of these research work has done the prediction by implementing pre-processing and feature reduction.

A few research studies have focused on implementing neural networks for agricultural commodity prediction [11] [12] [13]. However most of these work are about interval prediction and the point forecasting of agricultural commodity prediction has been taken less notice. Further on, because of gradient vanishing, these existing methods fail to capture very long-term information [11]. Next, the dynamic dependencies among multiple variables are not being taken into consideration [12]. Besides that, these studies fall short in distinguishing a mixture of short-term and long-term repeating patterns explicitly [13].

During the recent years, research papers have discussed the promising tools in time series forecasting provided by deep neural networks that can be classified into three categories [14]. First category is to identify statistically significant events, second is to find and predict inherent structure and third is to do accurate prediction on numerical value [15]. Looking into time series prediction through deep neural networks, there are several famous approaches which include Long Short Term Memory (LSTM) and Support Vector Machine (SVM). These approaches are highly discussed due to their result of their capability to not only model nonlinear patterns, realize complex causal relationships, as well as the learning rate on huge historical datasets. The LSTM model is said to be more accurate than the prediction of the ARIMA model by 85% on average from the results obtained in the research by Namini, S. S, Tavakoli, N. and Namin, A. S [16]. Furthermore, LSTM were introduced by and it aimed for a better performance by tackling the vanishing gradient issue that recurrent networks would suffer when dealing with long data sequences [17].

Other background information and technical review of ARIMA, SVR, Prophet, XGBoost and LSTM could be found from our conference paper [1].

## 3. Proposed System

The main objective of the proposed automated agriculture commodity price prediction system is to assist government or farmers for a better agricultural plantation plan. To achieve the objective, this research utilizes machine learning techniques to provide an agriculture price forecasting feature into the web system.

The project will study the needs of the farmers in doing agricultural activities, and these studies will be adapted into the system and be delivered in a simpler, more comprehensive way to suffice the farmers' knowledge in doing agricultural activities.

### 3.1. System Requirement Specification
### 3.1.1. Functional Requirements

The system should have a web app which consists of:

- A sign up and login page
- A forecast page to show all graph and data
- A commodity information page
- A user profile page
- An enquiry page

The system should forecast the future prices of agriculture commodities

- The forecast should be shown in a graph form
- User should be able to choose duration of prediction
- User should be able to rescale the x-axis of the graph
- The forecast should be based on previously available data
- User shall be able to access different type of model in price forecast (Univariate / multivariate)
- Graph should be updated upon type of commodity, duration of view, forecast period, and type of model

Visualization of forecast result should be clear for users

- Past prices and forecast results shall be separated with different color in one graph
- When the cursor hovers above the graph line, price information of the x-axis value shall be shown to users

Users shall able to select interested commodity

- Commodity chosen can only be pre-existing in the database
- System should update graph to show new data point based on new data added
- User shall be able to access different model type for each commodity (univariate, multivariate)

System should store and retrieve data in a database

- Developers should be able to update the database with new data
- Developers should be able to edit previous data stored in the database

Commodities data shall be downloadable

- User shall be able to download complete past prices of selected commodities in .csv file

A user profile shall be created automatically upon registration

- User shall be able to edit the user profile and save it

System shall be able to authenticate existing user

- User account shall be saved in database during registration

### 3.1.2. Non-Functional Requirements

Reliability

- The system shall be able to access the system anywhere with an internet connection
- The system shall be able to handle user queries not exceed 30 seconds when Tensorflow backend is running

Usability

- The system shall be simple to access for a first-time user
- The redirecting page amount for any feature of the system shall not exceed 5 pages

Performance

- The forecasting result shall be available for at least 7 commodities
- For each commodity, univariate model and multivariate model for forecast process shall be able to be accessed
- The meantime of download a file in csv format from a software shall not exceed 10 seconds

Security

- The system would require user account to access to all features
- User shall register their account with their email address one time only, system will reject account registration of existing account

Commodity products

- Commodities listed in the software shall be in English
- All commodities prices in the software shall be in Ringgit Malaysia, and local pricing for every commodities

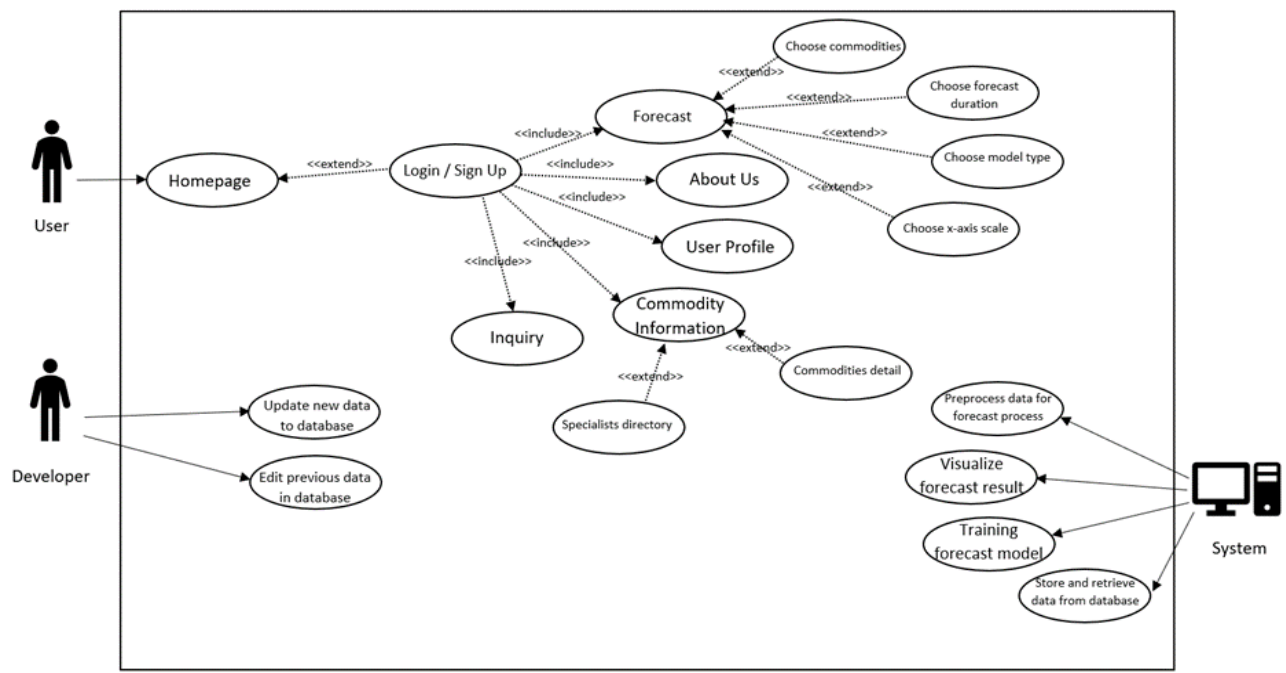

Figure 1 Use Case Diagram

Interface

- The system shall be portable in devices including personal computers, iPad and any mobile devices with Google Chrome software installed.
- The system layout shall be responsive to both full-size window and minimized window

*3.2. Use Case Diagram*

The use case diagram is shown in Figure 1. It explains the interactions that occur between the users and the system itself.

*3.3. System Overview*

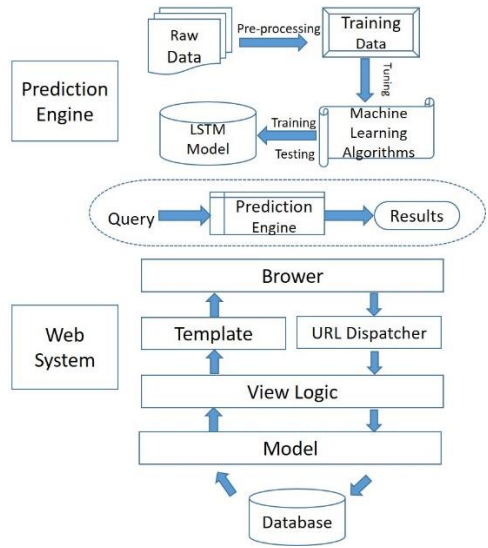

Figure 2 System Flow

Figure 2 describes the system Flow. The web system follows the Model-View-Controller (MVC) architecture. When a user enters a URL in their browser, the browser sends an http request to the web server, then the web server forwards the request to the application server and the URL setting contained in the urls.py file selects the view according to the url specified in the request, the view communicates with the database via models.py, renders the html or other format using templates (loads the static files) and returns the http response to the web server and finally, the web server provides the desired page to the browser.

*3.4. Prediction Engine Design*

There are four processes in the prediction engine design of the proposed system: data pre-processing process, tuning process, training and testing process and decision-making process. Process 1 involves cleaning and reformatting the dataset; while process 2 focuses on defining optimal parameter value of each machine learning techniques. All machine learning models will be trained and tested in process 3. Five machine learning methods were compared in this research, which are ARIMA [18], SVR [19], Prophet [20], XGBoost [21] and LSTM [17]. Technical details of these methods have been presented in our conference paper [1]. Finally, process 4 selects the best model out of all models to serve as the prediction engine.

*3.5. Implementation Practice*

The main implementation of the application for machine learning algorithms to predict, is Python programming language. It is an interpreted, high-level, general-purpose programming language used widely for machine learning. For the frontend design implementation of the application, Hypertext Markup Language (HTML), Cascading Style Sheets (CSS) and JavaScript have been used. These are high-level languages widely used to build a design and features of web applications.

*3.5.1. Collaborative Software and Version Control*

Django will be the main framework used to build our web application. It is Python-based software that follows the model-template-view architectural pattern. Besides, a free cloud service which supports free GPU named Google Colab is in service to assist us while implementing Long short-term memory (LSTM) algorithms in python programming language since it also supports the development for deep learning applications using popular libraries such as TensorFlow, Keras, PyTorch and so on. This helps us to share the identical up-to-date code with each other and each member is able to make changes to the code.

Github, an open source version control website, was used as a centralized repository to keep, share, update every member's codes and most importantly is to do version control. This platform is used as our service-oriented architecture for our application, also to manage the codebased, deliver the latest code, manage configuration of our web application.

*3.5.2. Graphical User Interface (GUI)*

During the initial implementation to achieve our prototype user interface (UI), we have used Hypertext Markup Language (HTML) to build the main structure of the homepage, login page and the forecasting dashboard. Thus, we modify using Cascading Style Sheet (CSS) and JavaScript to implement our proposed design accordingly. Throughout the continuous development, there are also quite some changes made to the frontend, these includes, the design of each page, the displayed functionality for relevant pages and the representation of the overall website. The programming languages used remain the same in the entire project to achieve the desired output for UI.

In the latest design, there are 8 pages that build up the website, particularly, register, login, dashboard, user profile, support, weather forecast, about us and subscription page. The following are supposed to be the user flow when interacting with the interface. Firstly, the user will be required to have an account in order to login the page, this will be handled in the register page for the user to create an account. Secondly, after getting an account, the user will be redirected to the login page with a simple input form for the user to log in with their account. All of the pages will require the user to login in order to view it. After logging in, there will be a side navigation bar containing each Uniform Resource Locator (URL) for relevant pages. In the forecast page as shown in Figure 3, users will be allowed to view the forecast result as a graph and users will also be able to select relevant commodities, data types, and duration of forecast they would prefer to see in the graph. Secondly, the user profile will handle all the information that belongs to the user and they are able to edit and update their profile. Thirdly, in the support page, users may submit their enquiries via the form provided. The about us page will be a brief introduction regarding the website and its purpose and usage. In the subscription page, the user will be able to subscribe to different plans to unlock different features. Lastly, each page contains the identical top navigation bar for them to log out anytime.

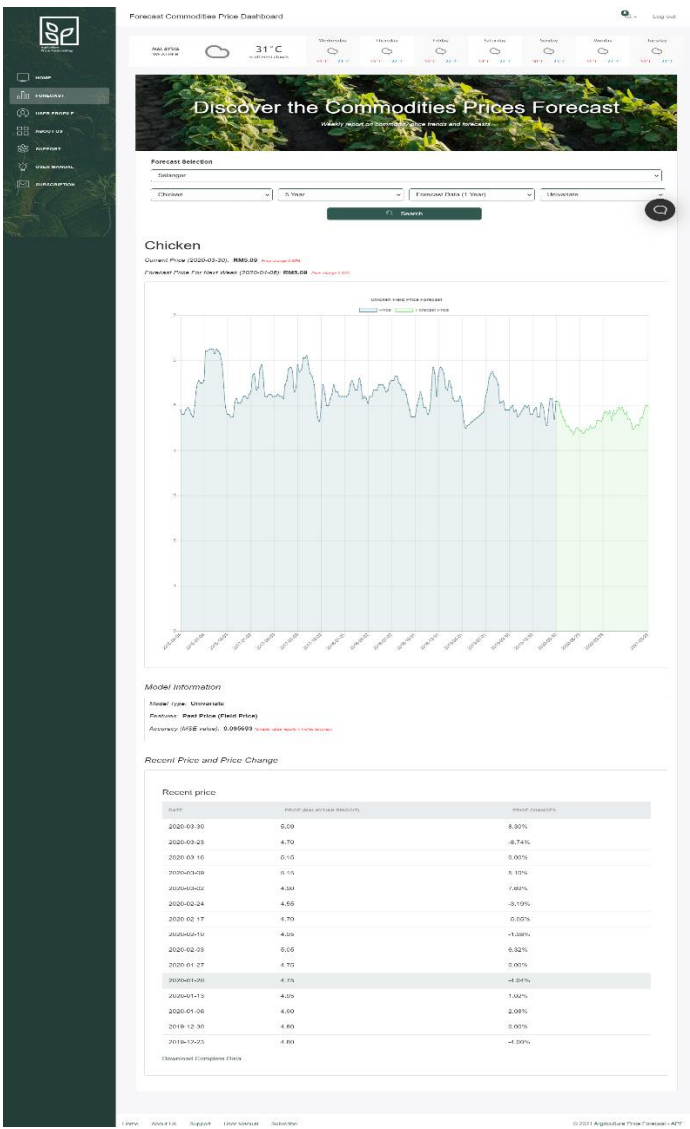

Figure 3 Forecasting Page

## 4. Experiments and Results

The forecasting feature of the software is the main focus of the project. In the decision of technique implemented to the software, an experiment has been designed to discover the most optimal algorithm which performs with high accuracy and capability of handling increasing data.

From the literature review, five algorithms namely, ARIMA, SVR, Prophet, XGBoost and Long short-term memory were selected as our potential prediction engine by comparing their performance to learn the best model of interest from large datasets.

*4.1. Datasets*

Our experimental agriculture datasets are extracted from the report made by FAMA official government website (https://sdvi.fama.gov.my/analisahargamingguan/DownloadPassword.asp). This website consists of weekly data analysis reports from 2007-2020. The price datasets of different categories (December 2008 to March 2020) are selected and migrated into a raw datasets. Two series of experiments have been conducted, one is to consider only time-series dataset with 11 years' datasets (December 2008 to March 2019) and the other is with multivariable (Temperature, Humidity, Precipitation and Crude Oil Price) with all datasets.

Table 1: Price Statistics of the Raw Data

| Commodities | Mean | Minimum | Maximum | StdDev |
|---|---|---|---|---|
| Chicken | 4.84 | 3.50 | 6.25 | 0.52 |
| Chili | 5.92 | 2.90 | 12 | 1.55 |
| Tomato | 2.19 | 0.50 | 6.35 | 0.83 |

Table 1 shows the statistics of the price with three selected commodities from two categories, chicken for poultry, chili and tomato for fruits. All three commodities datasets with 2% missing values.

*4.2. Experimental Setting*
*4.2.1. Configuration of ARIMA*

ARIMA is a combination of two models that is AR and MA. ARIMA consists of three parameters (p,d,q) where: p is the number of autoregressive terms, d is the number of nonseasonal differences needed for stationarity, and q is the number of lagged forecast errors in the prediction equation [18].

A series of actions has been performed to preprocess the data. Firstly, an Augmented Dickey Fuller Test (ADCF) is performed to determine whether the time series is stationary and perform data transformation to assure the stationarity of the data [22]. And then a differencing method is used to transform a non-stationary time series into a stationary one, a lag of 1 has been given as the result shows that the mean and variance are more constant over time in comparison to before transformation, thus, resulting in ARIMA (d,1).

In terms of auto regressive (AR), it demonstrates the correlation between the previous time period with the current. Thus, in order to predict the value of P, a partial auto-correlation function (PACF) graph will plot. This plot provides a summary of the relationship between an observation in a time series with observations at prior time steps with the relationships of intervening observations removed [23].

Inevitably, there is always noise or irregularity attached in a time series. Hence, in moving average (MA), the main purpose is to figure and average out the noises that could potentially affect the model. This can be achieved by plotting an ACF autocorrelation plot. An ACF plot is an (complete) auto-correlation function which gives values of auto-correlation of any series with its lagged values, it describes how well the present value of the series is related with its past values [22][24].

From our experimental results, it is observed that the diagram shows that the graph cuts off and drops to zero for the first time on x-axis in 1 on ACF plot and 1.5 on PACF plot, thus, giving q=1 and p=1.5.

*4.2.2. Configuration of SVR*

To capture the non-linear pattern of the data prediction, the implemented SVR model used Radial Basis Function as the kernel function. As a popular function used in most kernelling algorithm, it was claimed of having well performance under general smoothness assumptions [25] and this suits the situation in our study. All parameter settings remain default, however, some parameter values were specified in building the model, such as:

- The kernel coefficient (gamma) setting value is set as 1/total number of features to deal with various amounts of features.
- Regulation parameter(C) is set by fine-tuning to find the most suitable value for the model.
- The epsilon setting is used with default value.

To resolve the missing values, two different approaches are implemented to the model, such as:

- The missing value in the initial raw data - fill the missing value with valid, but not influence value (e.g. -99999) to avoid clearance of missing data affecting the prediction process.
- The missing value in the post-process raw data - clear the missing data to maintain the same amount of row in the data frame.

Training of the model utilized the python sklearn function called train_test_split. The dataset is randomly split into the corresponding train set and test set, in this case, the ratio of training set and test set is 9:1.

*4.2.3. Configuration of Prophet*

The default setting of Prophet:

- growth ('linear')
- n.changepoints (n)
- changepoints.range (0.8)
- changepoint_prior_scale (m)

The number of change points n has been set as one per month. The changepoint_prior_scale is to decide how flexible the changepoints are allowed. To avoid the overfitting problem, m has to set to [10,30]. Prophet is relatively robust to missing data, so no resampling methods have been implemented. Cross-validation method has been implemented to split the date into training and testing sets.

*4.2.4. Configuration of XGBoost*

The default setting of XGBoost:

- objective='reg:linear',
- colsample_bytree=0.8,
- learning_rate=0.1,
- max_depth=8,
- n_estimators=1000,
- silent=1,
- subsample=0.8,
- scale_pos_weight=1,
- seed=27
- early_stopping_rounds=50

*4.2.5. Configuration of LSTM*

The default setting of LSTM:

- Number of epochs: 100
- Batch size: 10
- Layers: 4 LSTM layer (including input layer) each of size 50 unit and subsequent by a dropout layer of +/-0.2 based on the dataset after each layer. 1 Output dense layer of size 52.
- Optimizer and loss function: adam, mse.
- Activation: hyperbolic tangent
- Data transform: data are scaled by a minmaxscaler

*4.3. Results*

Figure 4 represents the final mean squared error score (MSE) for the prediction in the first series of of experiments. The mean squared error score can suggest a better justification of the algorithm that had been chosen to implement this project, considering the advantages and disadvantages of each algorithm as well as other factors or parameters that need to be taken into account. From the results, ARIMA has shown as the most stable model amongst the three commodities, as it has the lowest mean square error value (0.251) on average, particularly with Chili, it gets as low as 0.027 for MSE.

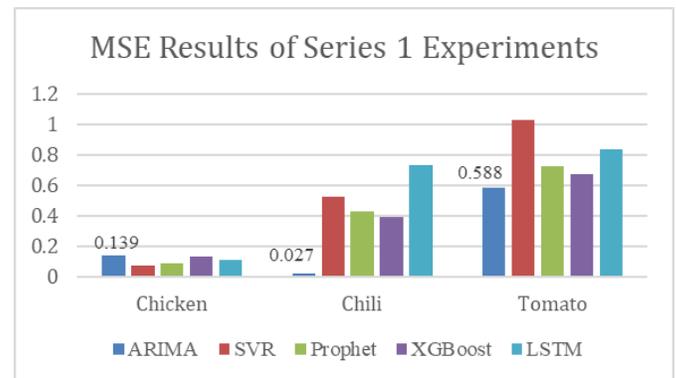

Figure 4 MSE Results of Series 1 Experiments

From the literature review , LSTM is more superior than other machine learning models such as SVR and ARIMA in terms of accuracy and difficulties in handling the data. However, due to the size of data which is too small for the model, it was not able to perform outstanding predictions to fit the system. On the other hand, ARIMA has outstanding ability in handling small amounts of data in performing predictions. Hence, from the result, ARIMA is proposed to be the most potential model which can be implemented to the software forecasting feature.

In testing the long-term potential prediction engine of the system, a second series of experiment have been done. In the second attempt of the experiment, the data has been updated to March 2020, and the feature space has been increased from one dimension to multi-dimensions with Temperature, Humidity, Precipitation and Crude Oil Price features being added to the model training process. Figure 5 shows the latest result of the experiment.

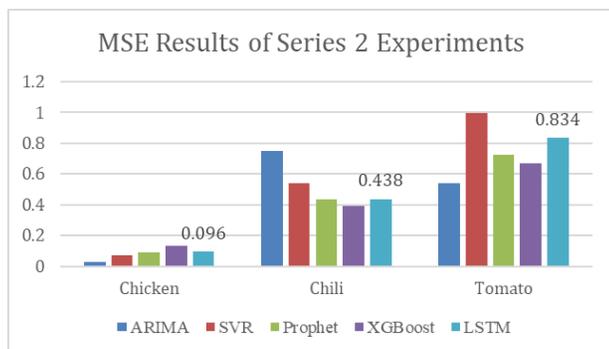

Figure 5 MSE Results of Series 2 Experiments

From the result, the average performance of LSTM with MSE has improved 45.5% while ARMIMA has dropped 74.1%. The average MSE of LSTM is 0.304 which outperformed all other four algorithms. The improvement of the LSTM model shows its potential in handling increasing data and higher complexity of data compared to other models and in a long-term overview, the collected data will continuously increase in terms of size and complexity, hence, LSTM is a better choice comparing to ARIMA (which shows the best result in the first series of experiment) due to its outperform capability of handling bigger dataset, as well as in the terms of system maintainability and scalability. Besides, this approach also provides room for parameters tuning according to the need.

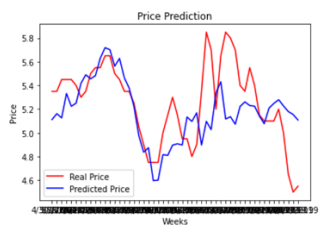

(a) Chicken Price Prediction

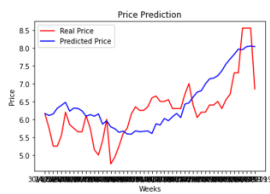 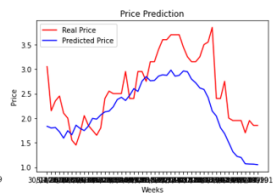

(b) Chili Price Prediction    (c) Tomato Price Prediction

Figure 6 LSTM Prediction Results

Figure 6 (a) to (c) are the graphs that illustrate the original prices, as well as the prediction prices using LSTM algorithm for 3 experimental commodities (Chicken, Chilli and tomato). The graph can help to visualise the outcome of predicted price in order to make a visual presentation to users.

## 5. Conclusion

In this research, we study and design an automated agriculture commodity price prediction system with novel machine learning techniques. Five popular machine learning algorithms, ARIMA, LSTM, SVR, Prophet and XGBoost have been compared in two series of experiments with large historical datasets in Malaysia and the most optimal algorithm, long short-term memory model has been selected as the prediction engine of the proposed system. The web-based system allows easy access to agriculture commodity price data and has the potential to facilitate government or farmers for a better plantation plan.

**Conflict of Interest**

The authors declare no conflict of interest.

**Acknowledgment**

The project idea is from FRGS Grant (FRGS/1/2018/ICT02/UNIM/02/1).